\documentclass{article} 
\usepackage{arxiv,times}

\usepackage{amsmath,amsfonts,bm}

\def\eqref#1{equation~\ref{#1}}

\def\1{\bm{1}}

\DeclareMathAlphabet{\mathsfit}{\encodingdefault}{\sfdefault}{m}{sl}
\SetMathAlphabet{\mathsfit}{bold}{\encodingdefault}{\sfdefault}{bx}{n}

\usepackage{hyperref}
\usepackage{url}
\usepackage{graphicx}
\usepackage{subcaption}
\usepackage{float}
\iclrfinalcopy

\title{VisCon-100K: Leveraging Contextual Web Data for Fine-tuning Vision Language Models}

\author{Gokul Karthik Kumar, Iheb Chaabane \& Kebin Wu \\
Technology Innovation Institute (TII)\\
Abu Dhabi, UAE\\
\texttt{\{gokul.kumar, iheb.chaabane, kebin.wu\}@tii.ae} \\
}

\begin{document}

\maketitle

\begin{abstract}
Vision-language models (VLMs) excel in various visual benchmarks but are often constrained by the lack of high-quality visual fine-tuning data. To address this challenge, we introduce VisCon-100K, a novel dataset derived from interleaved image-text web documents. Our approach transforms 45K web documents from the OBELICS dataset into 100K image conversation samples. We utilize GPT-4V to generate image-contextual captions and OpenChat 3.5 model to convert these captions into diverse free-form and multiple-choice question-answer pairs. Integrating this dataset for fine-tuning considerably enhances VLM performance across multiple benchmarks. Unlike methods that focus solely on fine-grained visual content, our approach leverages accompanying web context, yielding superior results. We also discover that a `leaky modality mix,' where conversation samples contain questions answerable from both the image and its contextual caption, outperforms non-leaky combinations of captions and Q\&A pairs. VisCon-100k dataset shows strong performance with two popular VLM approaches: text-only large language model (LLM) aligned with a vision encoder using image captions data (ShareGPT4V-7b) and multimodally pretrained LLM (IDEFICS2-8b) using interleaved image-text data. In addition to releasing the VisCon-100K dataset\footnote{\url{https://huggingface.co/datasets/tiiuae/viscon-100k}}, we provide a contextual captioner\footnote{\url{https://huggingface.co/tiiuae/viscon-contextual-captioner}} trained on this dataset, facilitating scalable fine-tuning data generation for future research and open-source applications. Using the same pipeline, but substituting our trained contextual captioner for GPT-4V, we also release the larger VisCon-1M dataset\footnote{\url{https://huggingface.co/datasets/tiiuae/viscon-1m}}.
\end{abstract}

\section{Introduction}

Recent advancements in large language models (LLMs) have revolutionized natural language processing (NLP), significantly impacting tasks such as text generation, summarization, translation, and question-answering. Models like LLaMA-2 \citep{touvron2023llama} and Mistral \citep{jiang2023mistral} have demonstrated exceptional capabilities, driving extensive research into their applications across various domains. Inspired by these successes, researchers have explored adapting LLMs for visual tasks, leading to significant developments in vision-language models (VLMs).

Two primary approaches have emerged for integrating visual understanding into LLMs:

\begin{enumerate}
\item \textbf{Alignment using Image Captions}: Popular models such as LLaVA-1.5 \citep{liu2024improved} and ShareGPT4V \citep{chen2023sharegpt4v} combine a pre-trained LLM with a CLIP \citep{radford2021learning}-based image encoder. The alignment of the image encoder’s output with the LLM is achieved through a two-stage training process: initially aligning the two modalities using image captions, followed by fine-tuning on vision-language tasks such as visual question answering (VQA).
\item \textbf{Multimodal Pretraining using Interleaved Image-Text}: These methods, including Kosmos-1 \citep{huang2024language} and IDEFICS2 \citep{laurenccon2024matters}, adopt a different strategy by performing multimodal pretraining. Using interleaved image-text web documents, they perform textual next-token prediction while incorporating visual context. This is typically followed by fine-tuning with VQA datasets.
\end{enumerate}

In addition to these two dominant approaches, several other methods such as Flamingo \citep{alayrac2022flamingo}, MiniGPT-4 \citep{zhu2023minigpt}, Prismer \citep{liu2023prismer}, Chameleon \citep{lu2024chameleon}, and Meta-Transformer \citep{zhang2023meta} adapt text-only LLMs for visual tasks. However, these alternatives, often involving more complex techniques, generally underperform on similar data and compute budgets compared to ShareGPT4V and IDEFICS2. Also, fine-tuning VLMs require considerable computational resources. So we evaluate our dataset and its design with the top representative models across two different popular VLM approaches: text-only large language models (LLM) aligned with a vision encoder using image captions data (ShareGPT4V-7b) and multimodally pretrained LLM (IDEFICS2-8b) using interleaved image-text data.

Despite these advancements, a critical gap persists: the scarcity of high-quality, diverse visual fine-tuning datasets. While extensive text-only fine-tuning datasets exist \citep{liu2024datasets}, there is a notable lack of vision-language datasets \citep{laurenccon2024matters} that provide the contextual richness required for effective vision-language understanding. Current datasets often fall short in capturing the broader web-based context that can enhance vision-language understanding.



To bridge this gap, we introduce \textbf{VisCon-100K}, a contextually rich dataset derived from interleaved image-text web documents. Our pipeline processes 45K web documents from the OBELICS \citep{laurenccon2024obelics} dataset into 100K image conversation samples. These samples are created by generating image-contextual captions using OpenAI GPT-4V API and transforming them into diverse free-form and multiple-choice question-answer pairs using OpenChat 3.5 \citep{wang2023openchat}. The resulting dataset, VisCon-100K, captures both \textbf{fine-grained visual descriptions} and \textbf{broader contextual information}, enabling more effective fine-tuning of VLMs.

\begin{figure*}[t]
  \centering
  \includegraphics[width=1\linewidth]{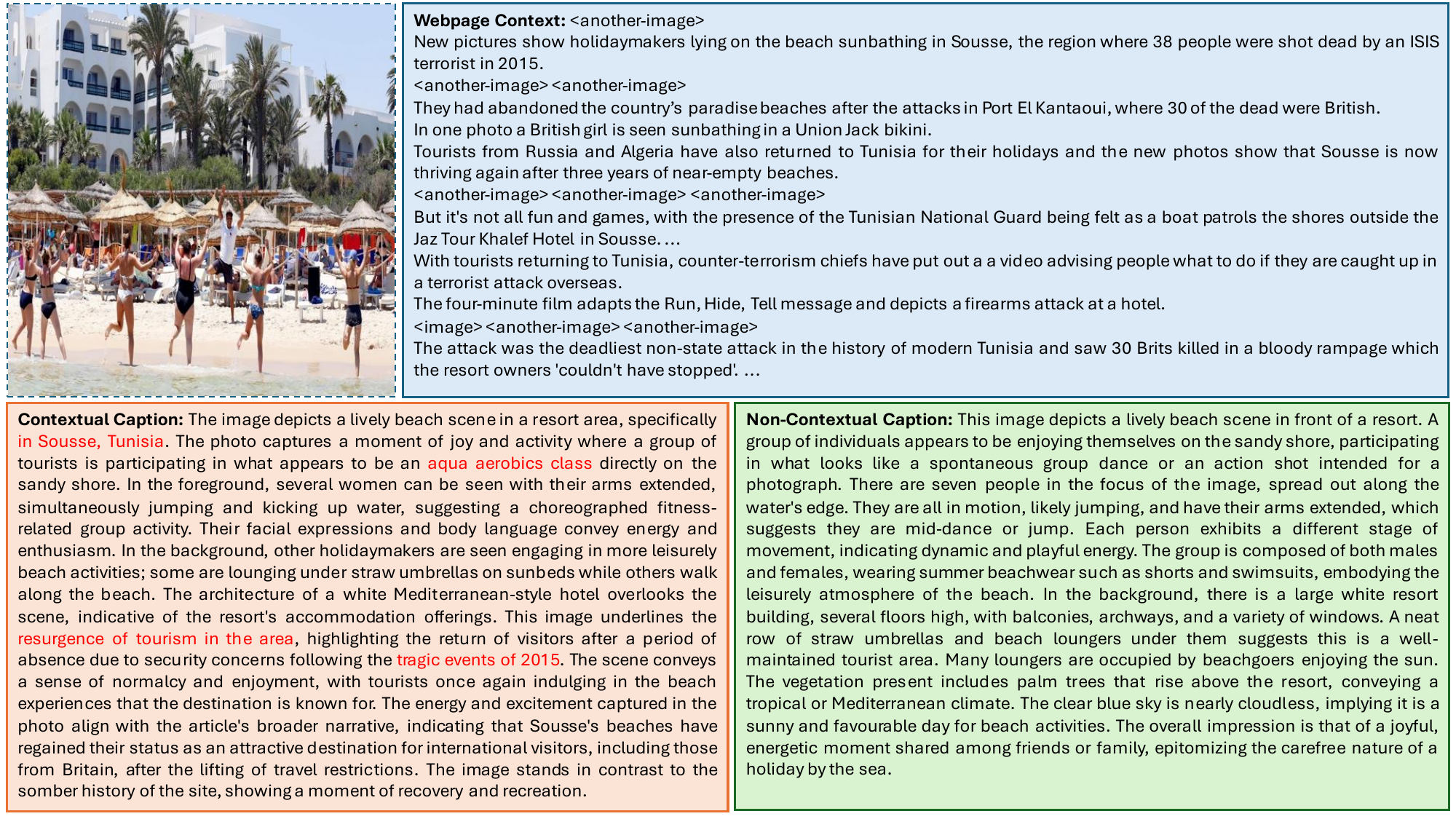}
  \caption{An OBELICS web document with generated contextual and non-contextual captions. The non-contextual caption describes the image in isolation, while the contextual caption integrates additional information from the surrounding web text, highlighted in red, providing a more nuanced and comprehensive description.}
  \label{fig:captions}
\end{figure*}

Our contributions can be summarized as follows:

\begin{enumerate}
    \item \textbf{Effective Use of Contextual Web Data}: We demonstrate the effectiveness of using contextual web data in combination with images, showcasing a sophisticated data generation pipeline that can be extended for future research and applications.
    \item \textbf{VisCon-100K Dataset}: We provide a novel, scalable dataset that notably enhances the performance of vision-language models across multiple benchmarks. By leveraging web context, VisCon-100K offers a richer and more diverse training resource than existing datasets.
    \item \textbf{Contextual Captioner}: We provide a trained contextual captioner to support scalable fine-tuning, enabling further research and open-source applications by generating high-quality contextual captions without relying on paid services like GPT-4V.
    \item \textbf{Leaky Modality Mix}: We introduce the concept of a "leaky modality mix," where conversation samples contain questions that can be answered from both the image and its contextual caption. This mix facilitates better integration of visual and textual information, outperforming non-leaky combinations of captions and Q\&A pairs.
\end{enumerate}
By addressing the need for high-quality visual fine-tuning data and demonstrating the benefits of incorporating contextual information, VisCon-100K represents a step forward in the development of robust vision-language models.

\section{Related Work}

Creating high-quality datasets for fine-tuning vision-language models is essential for improving their performance on complex multimodal tasks. Existing methods have made significant strides in this area, yet various challenges persist in terms of diversity, contextual richness, and scalability. Here, we discuss notable contributions and their limitations, setting the stage for the introduction of our approach used to develop \textbf{VisCon-100K}.

\paragraph{\textbf{Vision-Language Dataset Creation}}

\begin{enumerate}
    \item \textbf{Fine-Grained Image Captions}: Approaches such as those used in \textbf{ShareGPT4V} \citep{chen2023sharegpt4v}, \textbf{FuseCap} \citep{rotstein2023fusecap}, and \textbf{GranD} \citep{rasheed2024glamm} generate detailed image descriptions using LLMs. ShareGPT4V employs the GPT-4V API to produce detailed seed captions, aiming to reduce hallucinations and enhance dataset quality. Similarly, FuseCap integrates visual information from sources like object detectors and image taggers to enrich the captions, while GranD also queries LLM with a scene graph to add extra context. However, as these datasets scale, they tend to produce redundant descriptions of similar visual content, limiting their diversity and informativeness.
    \item \textbf{Contextual Data Utilization}: Some models, like \textbf{IDEFICS-2} \citep{laurenccon2024matters} and \textbf{Flamingo} \citep{alayrac2022flamingo}, employ contextual data in their pretraining by using interleaved image-text web documents. However, these approaches often retain a weak dependency on images while focusing on textual next-token prediction. The lack of grounding in the visual content means that the context derived from the web documents does not fully integrate with the image data, resulting in suboptimal alignment between visual and textual modalities.
    \item \textbf{Repurposing Classical Computer Vision Datasets}: Other methods, like \textbf{LLaVA} \citep{liu2024visual}, \textbf{ALLaVA} \citep{chen2024allava} and \textbf{IDEFICS-2} \citep{laurenccon2024matters}, attempt to repurpose datasets from common computer vision tasks for vision-language fine-tuning. While useful, these datasets often lack the diversity and contextual richness needed for real-life image conversations. They typically provide limited contextual information and fail to capture the broader web-based context that can enhance vision-language understanding. Moreover, these datasets often exhibit modality isolation, where questions are answerable either from a visual or a textual modality, but not both.

    
\end{enumerate}

\paragraph{\textbf{Challenges and Limitations}}

\begin{itemize}
    \item \textbf{Redundancy}: A common issue with current methods is the generation of redundant information, especially when scaling up the dataset. Repeated descriptions of similar content can reduce the dataset's overall effectiveness in training robust VLMs.
    \item \textbf{Lack of Contextual Grounding}: Many approaches show limited ability to generate data that is both contextually rich and relevant to real-life applications.
    \item \textbf{Modality Isolation}: Existing fine-tuning methods often treat visual and textual data separately, leading to a lack of integration between the two modalities. This isolation results in models that may excel in either visual understanding or textual comprehension but struggle to combine these insights effectively.
\end{itemize}



By conditioning image captioning on accompanying web content, \textbf{VisCon-100K} ensures the generated captions are \textbf{unique} and \textbf{contextually relevant} even as the dataset scales. This approach mitigates redundancy and enhances the dataset's relevance by leveraging the surrounding web context, thereby offering a more comprehensive training resource.
Figure~\ref{fig:captions} illustrates this approach, showing a web page containing an image along with its non-contextual and contextual captions. The non-contextual caption describes the image in isolation, while our contextual caption integrates relevant information from the surrounding web content, providing a more nuanced and comprehensive description. Furthermore, our adaptation of the \textbf{leaky modality mix} in conversations provides an opportunity for interplay between visual and textual modalities with their tighter integration potentially.

\begin{figure*}[t]
  \centering
  \includegraphics[width=1\linewidth]{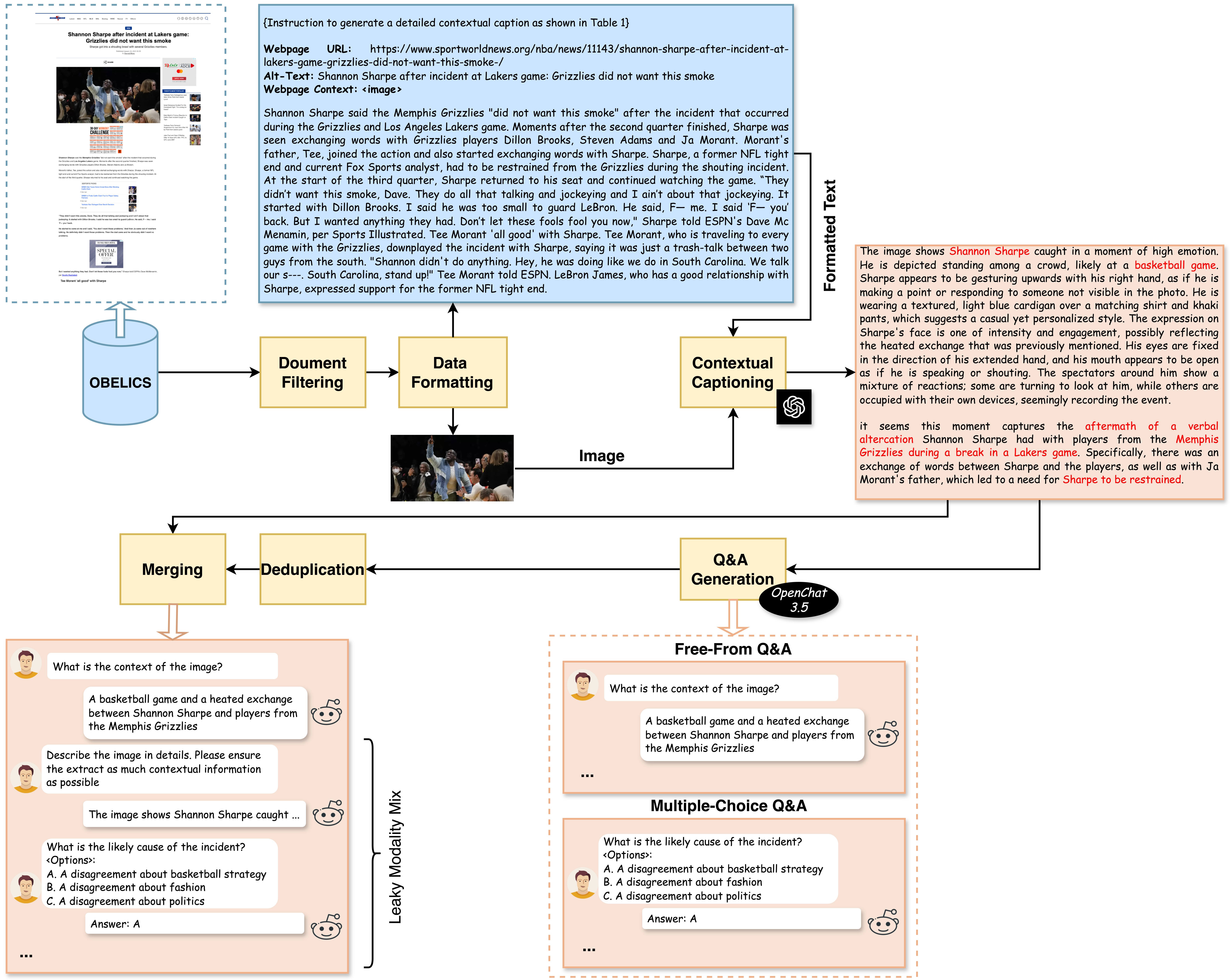}
  \caption{Data generation pipeline for creating the VisCon-100K dataset.}
  \label{fig:pipeline}
\end{figure*}

\begin{figure*}[]
  \centering
  \includegraphics[width=0.9\linewidth]{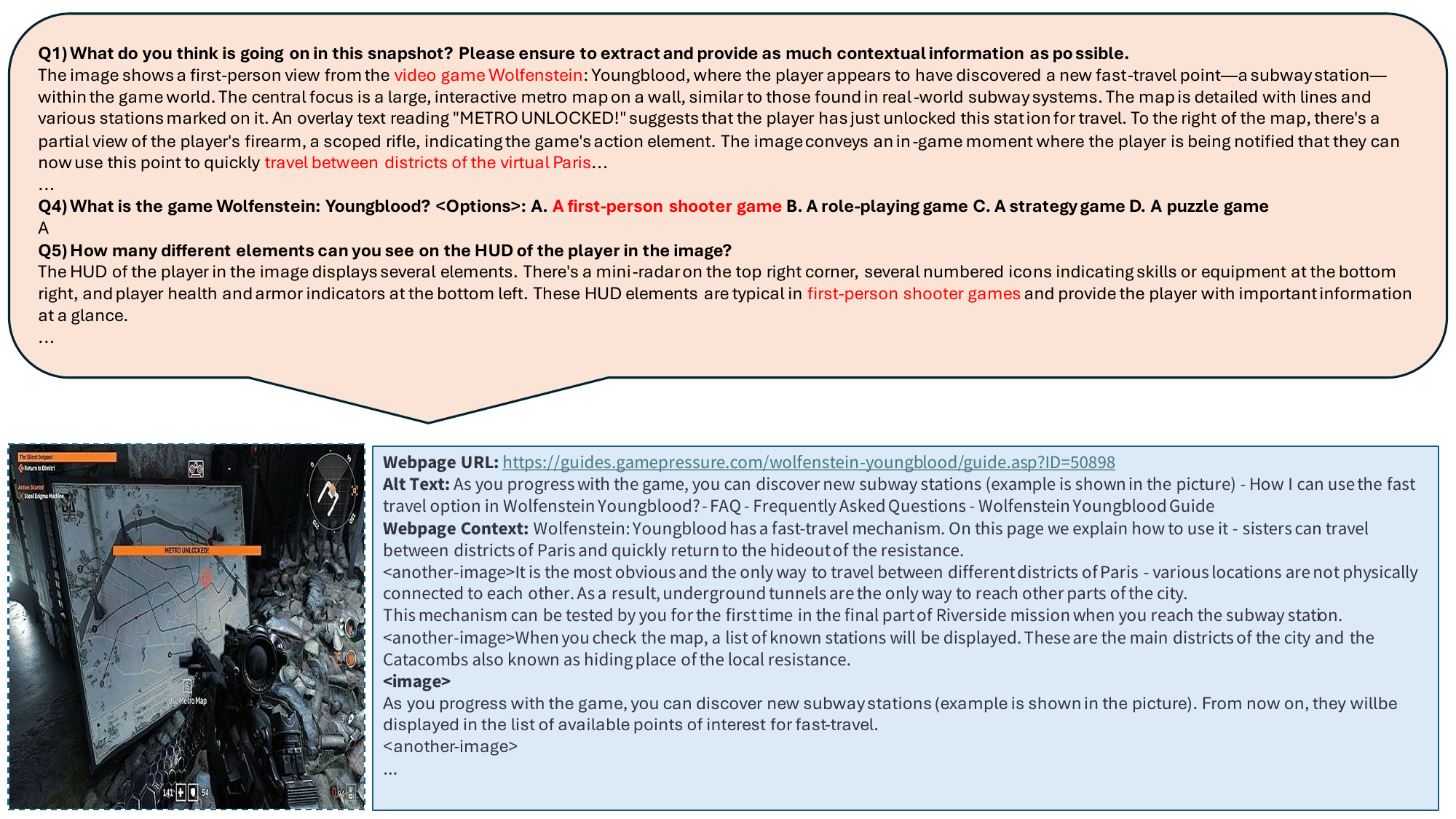}
  \includegraphics[width=0.9\linewidth]{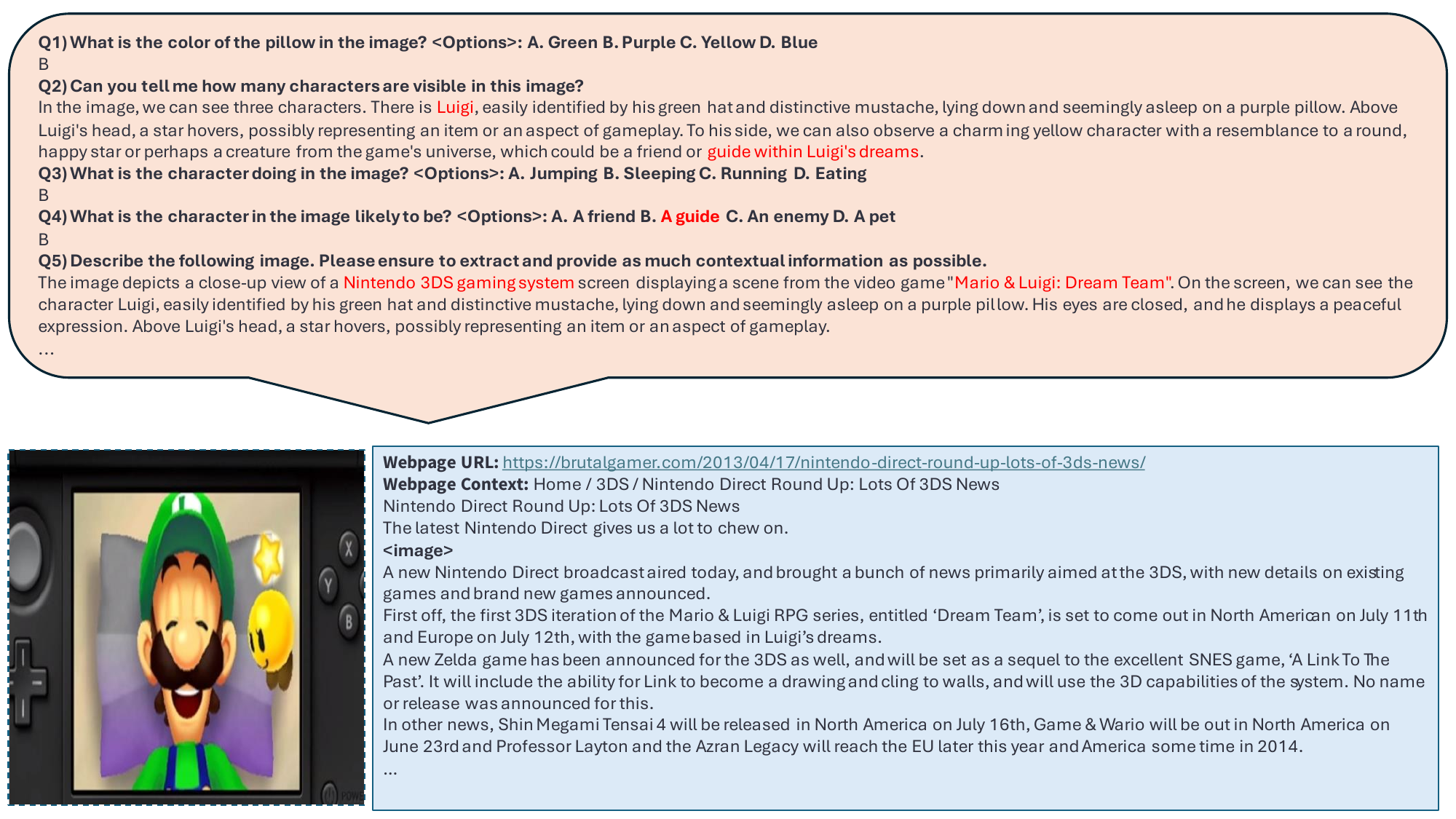}
  \includegraphics[width=0.9\linewidth]{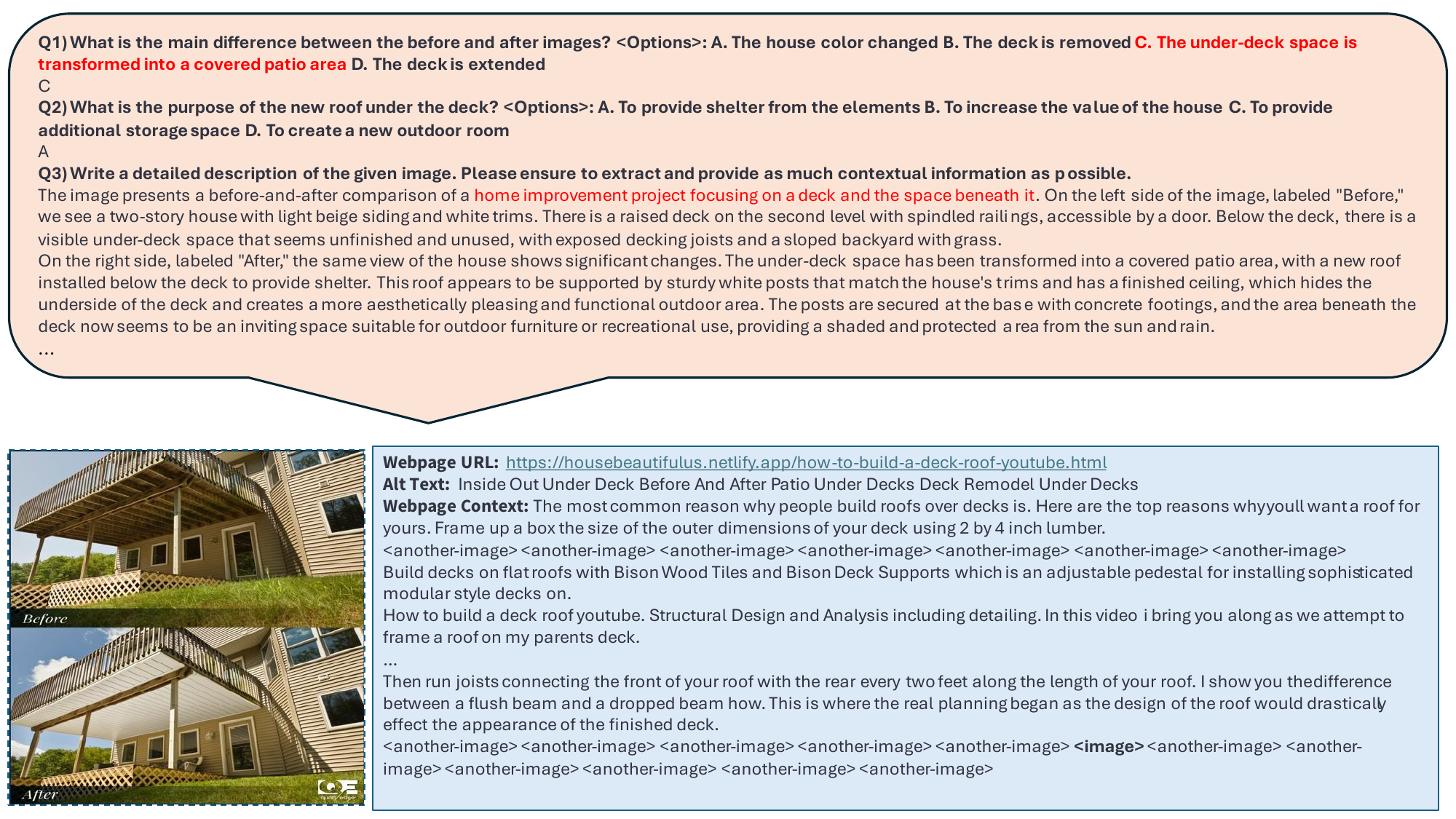}
  \caption{Examples from the VisCon-100K dataset. The text, highlighted in red, shows contextual grounding.}
  \label{fig:examples}
\end{figure*}

\section{Data Generation Pipeline}
\label{sec:data-pipeline}

Our approach leverages interleaved image-text web documents to generate, \textbf{VisCon-100K}, a contextually rich fine-tuning dataset for vision-language models (VLMs). The data generation pipeline involves several steps: document filtering, contextual captioning, Q\&A generation, deduplication and merging. The entire process is illustrated in Figure~\ref{fig:pipeline}. We show the example conversations from VisCon-100K in Figure~\ref{fig:examples} and analyze the properties of the dataset in Appendix~\ref{sec:dataprop}.

\subsection{Document Filtering}

We begin by filtering the OBELICS web documents to include only those with a maximum of 2000 text tokens, as determined by the Vicuna-7b \citep{zheng2024judging} tokenizer. This step ensures that each document provides sufficient context while remaining manageable in size. Notably, more than 90\% of the documents in OBELICS contain fewer than 2000 tokens.

\subsection{Contextual Captioning}

To generate contextual captions, we initially tested open-source VLMs like ShareGPT4V and LLaVA v1.5. However, we found that these models were not fine-tuned with web-contextual grounding datasets and often failed to include sufficient contextual information, sometimes even introducing hallucinations. In our qualitative evaluation with 100 samples, we discovered that GPT-4V significantly outperforms these models in producing high-quality contextual captions, especially when compared to non-contextual captions. Hence, we choose GPT4-V for this stage.


For each filtered web document, we extract relevant contextual information, including the webpage URL, image alt-text, and surrounding text. We also incorporate \texttt{<image>} and \texttt{<another-image>} placeholders to indicate the locations of the primary image and other images within the text. These elements collectively enhance the grounding of the captions, providing a rich context that helps in generating more fine-grained, accurate, and informative descriptions. Our approach was qualitatively validated, confirming its effectiveness. The prompt we adopted in using GPT-4V for generating contextual captions is shown in Table~\ref{tab:gpt4_prompt} in the Appendix.

\subsection{Q \& A Generation}

Following the generation of contextual captions, we explored various large language models (LLMs) for creating diverse free-form and multiple-choice question-answer pairs. After experimenting with LLaMA2-7b \citep{touvron2023llama} , Mistral \citep{jiang2023mistral}, Vicuna-7b \citep{zheng2024judging}, OpenChat 3.5 \citep{wang2023openchat}, and Gemma-7b \citep{team2024gemma} on 100 samples, we qualitatively chose OpenChat 3.5, a 7-billion-parameter LLM, for its superior performance in Q\&A generation.  For the Q\&A conversion, we found that open-source model like OpenChat 3.5 was sufficiently effective without the need to experiment with GPT-4V.

The Q\&A generation is guided by a prompt adapted from LLaVA \citep{liu2024visual} to convert captions into conversations, including few-shot examples for generating free-form question answers. We modified the instructions and few-shot examples also to generate multiple-choice questions. These prompts are shown in Tables~\ref{tab:qa_prompt_ffq} and ~\ref{tab:qa_prompt_mcq} in the Appendix. Additionally, we implemented post-processing steps, such as matching identifier names with regular expressions and checking for pairs, to filter out poorly formatted outputs.

Including Q\&A pairs is essential, especially when scaling the dataset. At 100K samples, VisCon-100K constitutes roughly 15\% of the overall fine-tuning data. As we scale beyond 1 million samples—given our source dataset OBELICS has 353 million images—the percentage of VisCon will be much higher. In such a scenario, the role of Q\&A becomes more crucial, as it reduces the model's bias towards always generating detailed responses irrespective of the question asked.

\subsection{Deduplication and Merging}

We merge the generated contextual captions, free-form, and multiple-choice question-answer pairs into coherent image conversations. Since captions do not inherently have an input prompt, we create a question for each caption using a randomly chosen LLaVA prompt for detailed image description and add the extra instruction \texttt{"Please ensure to extract and provide as much contextual information as possible.}"

Given the observed duplication between free-form and multiple-choice questions, we perform deduplication to avoid redundancy and ensure a balanced representation of question types. The deduplication process involves the following steps:
    \begin{itemize}
        \item \textbf{Generate Sentence Embeddings}: Encode the questions into embeddings using AnglE model \citep{li2023angle} to compute the cosine similarity matrix.
        \item \textbf{Select Unique Questions}: Iteratively select the most unique questions while maintaining a minimum count for each Q\&A type (free-form and multiple-choice) using similarity scores.
        \item \textbf{Shuffle Conversation Rounds}: Shuffle the conversation rounds to avoid pattern bias in the order of questions and answers.
    \end{itemize}

We include both captions and Q\&A pairs in each dataset sample, despite potential overlaps in information. We term this approach as a \textbf{`leaky modality mix'}. This method integrates questions that can be answered from both the image and the contextual caption within a single conversation sample, creating a controlled overlap or "leakage" of information across modalities. Our experiments in Section~\ref{sec:exp-leaky} show that this leaky modality mix performs better than non-leaky combinations of captions and Q\&A pairs. 

\section{Contextual Captioning Model}

To facilitate further extensions and reduce reliance on the paid GPT-4V service, we trained a contextual captioning model using the 100K contextual captions generated in our dataset. We fine-tuned IDEFICS2-8b, to accept both images and web content as input, enabling them to produce contextual captions. This additional fine-tuning with our dataset ensures that these models can generate high-quality contextual captions without the need for GPT-4V.


\section{Experiments}
\label{sec:experiements}

To evaluate the effectiveness of \textbf{VisCon-100K}, we conducted comprehensive experiments using two state-of-the-art vision-language models: ShareGPT4V-7b and IDEFICS2-8b. Our goal was to assess the impact of integrating VisCon-100K into existing fine-tuning datasets and to explore the performance benefits of the "leaky modality mix." 


We did not directly compare our dataset with other VQA datasets because VisCon-100K is designed to complement, not replace, existing datasets. Importantly, while most other datasets focus on detailed image descriptions, our dataset includes contextual knowledge that extends beyond the image but remains closely related. To the best of our knowledge, we are the first to incorporate large-scale contextual information into a VQA dataset for vision-language models.

Additionally, we evaluated our dataset against its non-contextual version derived from the same source with the same number of images. This approach aligns with methods used in other vision-language datasets like LLaVA and ShareGPT4V. This experimental comparison demonstrates the effectiveness of VisCon-100K, highlighting the value of adding contextual information to enhance performance in vision-language tasks.
\subsection{Setup}

For our experiments, we used the following setup:

\begin{itemize}
    \item \textbf{Models}: We utilized the pre-trained versions of ShareGPT4V-7b \citep{chen2023sharegpt4v} and IDEFICS2-8b \citep{laurenccon2024matters}. For ShareGPT4V-7b, we performed full fine-tuning, while for IDEFICS2-8b, we employed parameter-efficient fine-tuning as recommended. Notably, for IDEFICS2-8b, we omitted image splitting, focusing instead on demonstrating the effectiveness of our data pipeline rather than optimizing for peak performance. Except for this, we followed the hyperparameters used in their original papers.
    \item \textbf{Fine-Tuning Data}: The fine-tuning setup for these models followed similar procedures as outlined in their original works, using their respective publicly available fine-tuning datasets. We augmented these datasets with 100K samples from VisCon-100K, roughly constituting a 15\% increase in data volume.
    \item \textbf{Training Infrastructure}: We finetuned the models using AWS SageMaker instance of type ml.p4d.24xlarge, equipped with 8×40 GB A100 GPUs. This took a maximum of 12 hours for 1 epoch.
    \item \textbf{Framework}: Both models were trained using Hugging Face Transformers with DeepSpeed for optimization. 
\end{itemize}

\subsection{Evaluation Benchmarks}

We assessed model performance across six diverse vision-language benchmarks:

\begin{itemize}
    \item \textbf{SEED-Image} \citep{li2023seed}: Comprising 14,232 samples, this benchmark covers categories like instance attributes, identity, interaction, location, counting, scene understanding, spatial relations, text understanding, and visual reasoning.
    \item \textbf{MMBench} \citep{liu2023mmbench}: With 6,666 samples, it includes perception and reasoning subcategories, such as coarse and fine-grained perception and relational, attribute, and logical reasoning.
    \item \textbf{MMMU} \citep{yue2024mmmu}: Featuring 11,500 samples from fields like accounting, biology, chemistry, engineering, literature, medicine, physics, psychology, and more.
    \item \textbf{AI2D} \citep{kembhavi2016diagram}: Includes 5,000 images with three questions per image, covering various academic topics.
    \item \textbf{ScienceQA} \citep{lu2022learn}: Consists of 2,000 samples across topics like astronomy, biology, geography, history, and physics.
    \item \textbf{LLaVA Bench} \citep{liu2024visual}: Contains 24 images with 60 questions focusing on visual conversation, detailed image descriptions, and complex visual reasoning. For scoring the answers, we used LLaMA3-8b for cost efficiency instead of GPT-4, comparing generated answers to reference texts.
\end{itemize}

\begin{table*}[t]
\caption{Performance of ShareGPT4V-7b model for different configurations on the SEED benchmark.}
\label{tab:results}
\centering
\begin{tabular}{p{0.55\textwidth} c}
\hline
\textbf{Configuration} & \textbf{Score} \\ \hline
Base Model (without contextual data) & 66.24 \\ \hline
\textit{Isolated Variants:} \\
+ Contextual Captions Alone & 66.9 \\ 
+ Free-form Q\&A Alone & 65.26 \\ 
+ Multiple-choice Q\&A Alone & 63.97 \\ \hline
\textit{Non-Leaky Mix Variants:} \\
+ Combination of Free-form and Multiple-choice Q\&A & 61.25 \\ 
+ Separated Samples of Captions and Q\&A  & 59.31 \\ \hline
\textit{Leaky Modality Mix:} \\
+ Combined Mix of Captions and Q\&A & \textbf{67.62} \\ \hline

\end{tabular}
\end{table*}



\subsection{Evaluating Data Combinations: The Impact of Leaky Modality Mix}
\label{sec:exp-leaky}

To determine the optimal data composition, we evaluated different configurations of VisCon-100K using the SEED benchmark with the ShareGPT4V-7b model. This step was crucial to identify the best approach for integrating captions and Q\&A pairs. We experimented with the following configurations:

\begin{itemize}
    \item \textbf{Contextual Captions Alone}: Using only the contextual captions.
    \item \textbf{Free-form Q\&A Alone}: Incorporating only the derived free-form question-answer pairs.
    \item \textbf{Multiple-choice Q\&A Alone}: Using only the multiple-choice question-answer pairs.
    \item \textbf{Combination of Free-form and Multiple-choice Q\&A}: Integrating both types of Q\&A pairs in each conversation but no captions.
    \item \textbf{Separated Samples}: Using one conversation sample for captions and another for Q\&A pairs. 
    \item \textbf{Combined Mix}: Incorporating a mix of all three (contextual captions, free-form Q\&A, and multiple-choice Q\&A) in each sample.
\end{itemize}

The performance for each configuration is shown in Table~\ref{tab:results}. Our results show that the \textbf{`leaky modality mix'}—a configuration where each sample includes questions that can be answered from both the image and its contextual caption—outperforms using captions or Q\&A pairs exclusively. This mix mitigates biases seen in configurations using only captions (which tend to generate lengthy descriptions) or only Q\&A pairs (which can overlook significant details). Additionally, by including both sources of information within a single conversation, the model can leverage the interplay between visual and textual data more effectively, leading to better integration and improved performance.


We also tested non-leaky mix configurations where captions were removed entirely or where captions and Q\&A pairs were split into different samples, to understand the impact of explicit information leakage. The findings indicate that controlled leakage across modalities enhances the model's ability to integrate visual and textual information, thereby improving overall performance. Although the improvement from the leaky modality mix over using contextual captions alone appears modest, statistical tests confirm its significance. McNemar's test between the leaky modality mix and the base model yields a p-value of $2.118 \times 10^{-5}$, and between the leaky modality mix and the contextual captions model, a p-value of 0.027—both indicating strong statistical significance.

\begin{figure}[t]
  \centering
  \begin{minipage}{0.49\columnwidth}
    \includegraphics[width=\linewidth]{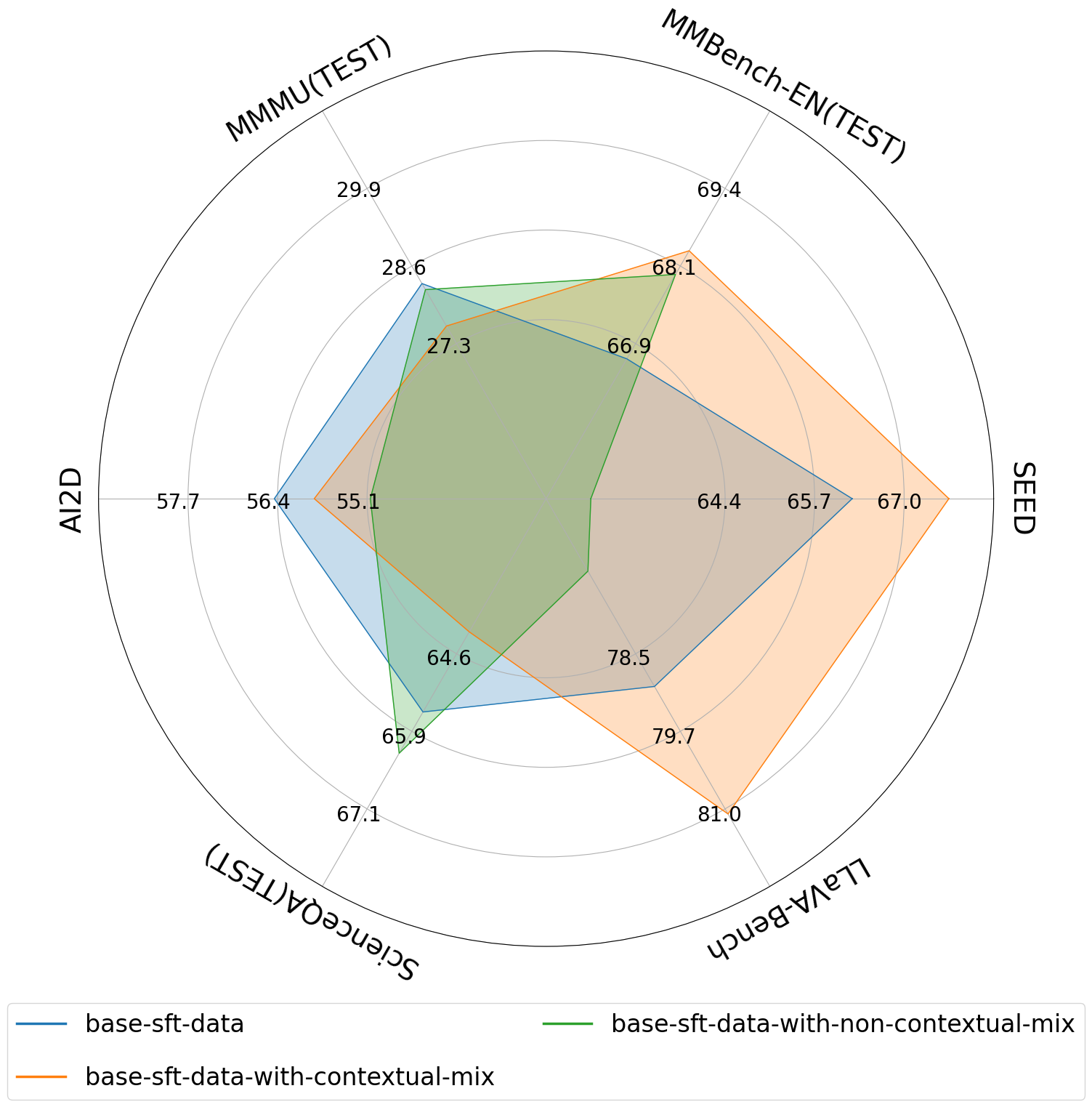}
    \caption{Performance of ShareGPT4V-7b model across 6 benchmarks for different data configurations}
    \label{fig:sharegpt4v-results}
  \end{minipage}\hfill
  \begin{minipage}{0.49\columnwidth}
    \includegraphics[width=\linewidth]{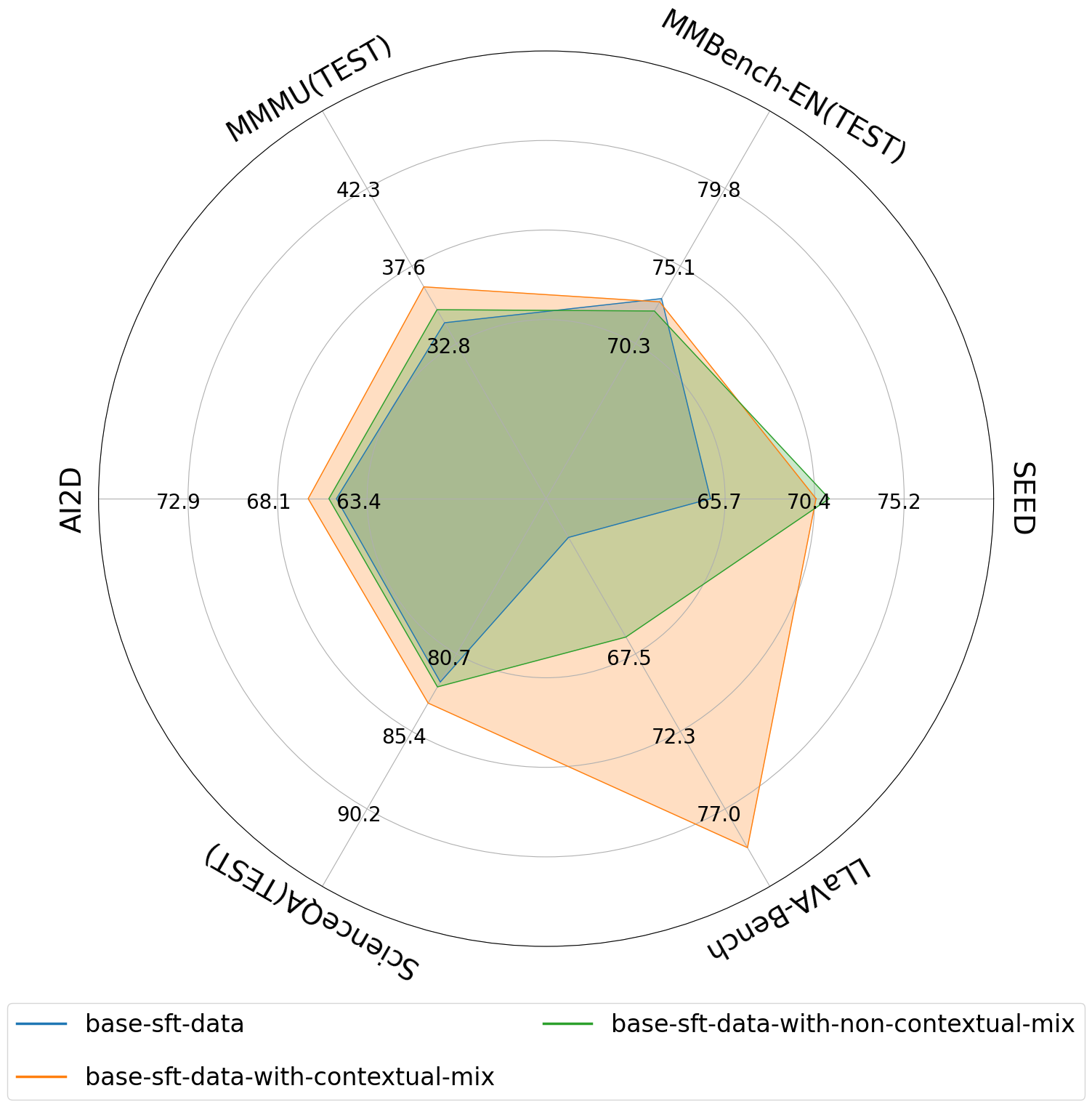}
    \caption{Performance of IDEFICS2-8b model across 6 benchmarks for different data configurations}
    \label{fig:idefics2-results}
  \end{minipage}
\end{figure}

\subsection{Contextual vs. Non-Contextual Data}

To get the non-contextual data, we followed the same pipeline described in Section~\ref{sec:data-pipeline}, but without incorporating the webpage context during captioning and using prompts adapted accordingly.

Using the optimal \textbf{leaky modality mix}, we extended our evaluation across all six benchmarks. The results, depicted in Figure~\ref{fig:sharegpt4v-results}, demonstrate that the contextual mix outperforms in 3 out of 6 benchmarks. Specifically, the contextual mix significantly boosts performance on SEED and LLaVA Bench. On average, across all benchmarks, the contextual mix scored the highest with an \textbf{average of 60.81}, followed by the base model at 60.35, and the non-contextual mix at 59.51.

Interestingly, adding non-contextual data did not provide any substantial benefit on average, likely due to redundancy with the base fine-tuning data. This observation suggests that contextual information is crucial for enhancing the dataset’s utility in vision-language tasks.

\subsection{Generalizability}

To assess the generalizability of our findings, we replicated the experiments with the more recent IDEFICS2-8b model, which is pretrained using interleaved image-text web documents, including OBELICS. Despite deriving our additional fine-tuning data from OBELICS itself, the contextual mix further improved IDEFICS2-8b’s performance, as shown in Figure~\ref{fig:idefics2-results}. In detail, across all benchmarks, the contextual mix scored an \textbf{average of 68.21}, compared to 65.50 for the non-contextual mix and 63.31 for the base model. In addition, the boost across different benchmarks is more consistent compared to the experiments with ShareGPT4V-7b. We attribute this to the stronger integration of image and text data provided by our contextual conversations.

Preliminary experiments were conducted with ShareGPT4V-7b as IDEFICS2-8b was not available during the initial stages of our research. Given the significant computational resources required for fine-tuning and evaluation, we focused subsequent tests on the most promising configurations. The consistent performance improvements with IDEFICS2-8b underscore the utility of VisCon-100K, suggesting potential for further enhancements by processing additional web documents at scale.

It is crucial to note that the performance improvement is not solely due to the increased dataset size but also due to our approach. Table~\ref{tab:results} shows that adding the same number of samples in non-leaky or isolated variants resulted in poorer performance. Figures ~\ref{fig:sharegpt4v-results} and ~\ref{fig:idefics2-results} further illustrate that adding contextual samples yields better results than adding non-contextual counterparts across multiple benchmarks, highlighting the impact of contextual information.

\subsection{Contextual Captioner}

To facilitate further extensions of VisCon-100K, we finetuned IDEFICS2-8b model using the 100K contextual captions in our dataset. Evaluations on a held-out set of 1894 GPT-4 generated contextual captions showed an \textbf{increase of 4 BLEU points} and \textbf{3 ROUGE-L F1 points} with finetuning.


\section{Conclusion}

In this work, we introduced \textbf{VisCon-100K}, a novel dataset derived from interleaved image-text web documents, designed to enhance the fine-tuning of vision-language models (VLMs). Our approach generates contextually rich image conversations by creating image-contextual captions and transforming them into diverse question-answer pairs. Experiments demonstrate that integrating VisCon-100K notably improves VLM performance across multiple benchmarks. Additionally, our \textbf{leaky modality mix} strategy enhances the interplay between visual and textual modalities. We also provide a contextual captioner to facilitate the scalable extension of VisCon-100K, supporting open-source research and applications.

\section{Future Work}
\label{future-work}

\begin{enumerate}
    \item \textbf{Multilingual Contexts and Scaling}: Extend the dataset to include multilingual web content, improving the generalizability and applicability of VLMs across different languages and cultural contexts. Additionally, scale the dataset to potentially over 300 million images, leveraging the full scope of the OBELICS dataset to enhance the depth and diversity of the fine-tuning data.
    \item \textbf{Expanding Data Types for Fine-tuning}: Incorporate more complex conversation types, such as dialogues involving multiple images or more intricate Q\&A formats, supported by ablation studies to determine the optimal mix of data types.
    \item \textbf{Advanced Post-Processing Techniques}: Develop sophisticated post-processing methods to ensure the uniqueness, harmlessness, and usefulness of the generated data, enhancing the dataset's reliability and safety.
    \item \textbf{Creating Diverse Benchmarks}: Establish comprehensive benchmarks to evaluate models on contextual visual question answering tasks, ensuring robust and generalizable model performance across varied scenarios.

\end{enumerate}

\section{Limitations}

Despite the promising results, our approach has some limitations:

\begin{enumerate}
    \item \textbf{Potentially Harmful Content}: While web data offers diverse contexts, it may include harmful or inappropriate content that our current pipeline does not explicitly filter out. Future work should incorporate robust content moderation techniques to mitigate these risks.
    \item \textbf{Reliance on GPT-4}: The use of GPT-4 for generating seed contextual captions provides a high-quality foundation for our dataset. However, GPT-4's performance in non-English languages and its reliance as a paid service may limit accessibility and introduce language biases. Our contextual captioner partially aims to address this by providing an open-source alternative, but further refinement is needed for broader applicability in multiple languages.
    \item \textbf{Quality of Contextual Information}: The quality and relevance of the contextual information extracted from web documents can vary significantly, potentially affecting the consistency and effectiveness of the fine-tuning data. Ensuring high-quality context extraction remains a challenge that requires continuous improvement.

\end{enumerate}


\bibliography{iclr2025_conference}

\begin{thebibliography}{27}
\providecommand{\natexlab}[1]{#1}
\providecommand{\url}[1]{\texttt{#1}}
\expandafter\ifx\csname urlstyle\endcsname\relax
  \providecommand{\doi}[1]{doi: #1}\else
  \providecommand{\doi}{doi: \begingroup \urlstyle{rm}\Url}\fi

\bibitem[Alayrac et~al.(2022)Alayrac, Donahue, Luc, Miech, Barr, Hasson, Lenc, Mensch, Millican, Reynolds, et~al.]{alayrac2022flamingo}
Jean-Baptiste Alayrac, Jeff Donahue, Pauline Luc, Antoine Miech, Iain Barr, Yana Hasson, Karel Lenc, Arthur Mensch, Katherine Millican, Malcolm Reynolds, et~al.
\newblock Flamingo: a visual language model for few-shot learning.
\newblock \emph{Advances in neural information processing systems}, 35:\penalty0 23716--23736, 2022.

\bibitem[Chen et~al.(2024)Chen, Chen, Zhang, Chen, Wu, Zhang, Chen, Li, Wan, and Wang]{chen2024allava}
Guiming~Hardy Chen, Shunian Chen, Ruifei Zhang, Junying Chen, Xiangbo Wu, Zhiyi Zhang, Zhihong Chen, Jianquan Li, Xiang Wan, and Benyou Wang.
\newblock Allava: Harnessing gpt4v-synthesized data for a lite vision-language model.
\newblock \emph{arXiv preprint arXiv:2402.11684}, 2024.

\bibitem[Chen et~al.(2023)Chen, Li, Dong, Zhang, He, Wang, Zhao, and Lin]{chen2023sharegpt4v}
Lin Chen, Jisong Li, Xiaoyi Dong, Pan Zhang, Conghui He, Jiaqi Wang, Feng Zhao, and Dahua Lin.
\newblock Sharegpt4v: Improving large multi-modal models with better captions.
\newblock \emph{arXiv preprint arXiv:2311.12793}, 2023.

\bibitem[Huang et~al.(2024)Huang, Dong, Wang, Hao, Singhal, Ma, Lv, Cui, Mohammed, Patra, et~al.]{huang2024language}
Shaohan Huang, Li~Dong, Wenhui Wang, Yaru Hao, Saksham Singhal, Shuming Ma, Tengchao Lv, Lei Cui, Owais~Khan Mohammed, Barun Patra, et~al.
\newblock Language is not all you need: Aligning perception with language models.
\newblock \emph{Advances in Neural Information Processing Systems}, 36, 2024.

\bibitem[Jiang et~al.(2023)Jiang, Sablayrolles, Mensch, Bamford, Chaplot, Casas, Bressand, Lengyel, Lample, Saulnier, et~al.]{jiang2023mistral}
Albert~Q Jiang, Alexandre Sablayrolles, Arthur Mensch, Chris Bamford, Devendra~Singh Chaplot, Diego de~las Casas, Florian Bressand, Gianna Lengyel, Guillaume Lample, Lucile Saulnier, et~al.
\newblock Mistral 7b.
\newblock \emph{arXiv preprint arXiv:2310.06825}, 2023.

\bibitem[Kembhavi et~al.(2016)Kembhavi, Salvato, Kolve, Seo, Hajishirzi, and Farhadi]{kembhavi2016diagram}
Aniruddha Kembhavi, Mike Salvato, Eric Kolve, Minjoon Seo, Hannaneh Hajishirzi, and Ali Farhadi.
\newblock A diagram is worth a dozen images.
\newblock In \emph{Computer Vision--ECCV 2016: 14th European Conference, Amsterdam, The Netherlands, October 11--14, 2016, Proceedings, Part IV 14}, pp.\  235--251. Springer, 2016.

\bibitem[Lauren{\c{c}}on et~al.(2024{\natexlab{a}})Lauren{\c{c}}on, Saulnier, Tronchon, Bekman, Singh, Lozhkov, Wang, Karamcheti, Rush, Kiela, et~al.]{laurenccon2024obelics}
Hugo Lauren{\c{c}}on, Lucile Saulnier, L{\'e}o Tronchon, Stas Bekman, Amanpreet Singh, Anton Lozhkov, Thomas Wang, Siddharth Karamcheti, Alexander Rush, Douwe Kiela, et~al.
\newblock Obelics: An open web-scale filtered dataset of interleaved image-text documents.
\newblock \emph{Advances in Neural Information Processing Systems}, 36, 2024{\natexlab{a}}.

\bibitem[Lauren{\c{c}}on et~al.(2024{\natexlab{b}})Lauren{\c{c}}on, Tronchon, Cord, and Sanh]{laurenccon2024matters}
Hugo Lauren{\c{c}}on, L{\'e}o Tronchon, Matthieu Cord, and Victor Sanh.
\newblock What matters when building vision-language models?
\newblock \emph{arXiv preprint arXiv:2405.02246}, 2024{\natexlab{b}}.

\bibitem[Li et~al.(2023)Li, Wang, Wang, Ge, Ge, and Shan]{li2023seed}
Bohao Li, Rui Wang, Guangzhi Wang, Yuying Ge, Yixiao Ge, and Ying Shan.
\newblock Seed-bench: Benchmarking multimodal llms with generative comprehension.
\newblock \emph{arXiv preprint arXiv:2307.16125}, 2023.

\bibitem[Li \& Li(2023)Li and Li]{li2023angle}
Xianming Li and Jing Li.
\newblock Angle-optimized text embeddings.
\newblock \emph{arXiv preprint arXiv:2309.12871}, 2023.

\bibitem[Liu et~al.(2024{\natexlab{a}})Liu, Li, Li, and Lee]{liu2024improved}
Haotian Liu, Chunyuan Li, Yuheng Li, and Yong~Jae Lee.
\newblock Improved baselines with visual instruction tuning.
\newblock In \emph{Proceedings of the IEEE/CVF Conference on Computer Vision and Pattern Recognition}, pp.\  26296--26306, 2024{\natexlab{a}}.

\bibitem[Liu et~al.(2024{\natexlab{b}})Liu, Li, Wu, and Lee]{liu2024visual}
Haotian Liu, Chunyuan Li, Qingyang Wu, and Yong~Jae Lee.
\newblock Visual instruction tuning.
\newblock \emph{Advances in neural information processing systems}, 36, 2024{\natexlab{b}}.

\bibitem[Liu et~al.(2023{\natexlab{a}})Liu, Fan, Johns, Yu, Xiao, and Anandkumar]{liu2023prismer}
Shikun Liu, Linxi Fan, Edward Johns, Zhiding Yu, Chaowei Xiao, and Anima Anandkumar.
\newblock Prismer: A vision-language model with multi-task experts.
\newblock \emph{arXiv preprint arXiv:2303.02506}, 2023{\natexlab{a}}.

\bibitem[Liu et~al.(2024{\natexlab{c}})Liu, Cao, Liu, Ding, and Jin]{liu2024datasets}
Yang Liu, Jiahuan Cao, Chongyu Liu, Kai Ding, and Lianwen Jin.
\newblock Datasets for large language models: A comprehensive survey.
\newblock \emph{arXiv preprint arXiv:2402.18041}, 2024{\natexlab{c}}.

\bibitem[Liu et~al.(2023{\natexlab{b}})Liu, Duan, Zhang, Li, Zhang, Zhao, Yuan, Wang, He, Liu, et~al.]{liu2023mmbench}
Yuan Liu, Haodong Duan, Yuanhan Zhang, Bo~Li, Songyang Zhang, Wangbo Zhao, Yike Yuan, Jiaqi Wang, Conghui He, Ziwei Liu, et~al.
\newblock Mmbench: Is your multi-modal model an all-around player?
\newblock \emph{arXiv preprint arXiv:2307.06281}, 2023{\natexlab{b}}.

\bibitem[Lu et~al.(2022)Lu, Mishra, Xia, Qiu, Chang, Zhu, Tafjord, Clark, and Kalyan]{lu2022learn}
Pan Lu, Swaroop Mishra, Tanglin Xia, Liang Qiu, Kai-Wei Chang, Song-Chun Zhu, Oyvind Tafjord, Peter Clark, and Ashwin Kalyan.
\newblock Learn to explain: Multimodal reasoning via thought chains for science question answering.
\newblock \emph{Advances in Neural Information Processing Systems}, 35:\penalty0 2507--2521, 2022.

\bibitem[Lu et~al.(2024)Lu, Peng, Cheng, Galley, Chang, Wu, Zhu, and Gao]{lu2024chameleon}
Pan Lu, Baolin Peng, Hao Cheng, Michel Galley, Kai-Wei Chang, Ying~Nian Wu, Song-Chun Zhu, and Jianfeng Gao.
\newblock Chameleon: Plug-and-play compositional reasoning with large language models.
\newblock \emph{Advances in Neural Information Processing Systems}, 36, 2024.

\bibitem[Radford et~al.(2021)Radford, Kim, Hallacy, Ramesh, Goh, Agarwal, Sastry, Askell, Mishkin, Clark, et~al.]{radford2021learning}
Alec Radford, Jong~Wook Kim, Chris Hallacy, Aditya Ramesh, Gabriel Goh, Sandhini Agarwal, Girish Sastry, Amanda Askell, Pamela Mishkin, Jack Clark, et~al.
\newblock Learning transferable visual models from natural language supervision.
\newblock In \emph{International conference on machine learning}, pp.\  8748--8763. PMLR, 2021.

\bibitem[Rasheed et~al.(2024)Rasheed, Maaz, Shaji, Shaker, Khan, Cholakkal, Anwer, Xing, Yang, and Khan]{rasheed2024glamm}
Hanoona Rasheed, Muhammad Maaz, Sahal Shaji, Abdelrahman Shaker, Salman Khan, Hisham Cholakkal, Rao~M Anwer, Eric Xing, Ming-Hsuan Yang, and Fahad~S Khan.
\newblock Glamm: Pixel grounding large multimodal model.
\newblock In \emph{Proceedings of the IEEE/CVF Conference on Computer Vision and Pattern Recognition}, pp.\  13009--13018, 2024.

\bibitem[Rotstein et~al.(2023)Rotstein, Bensaid, Brody, Ganz, and Kimmel]{rotstein2023fusecap}
Noam Rotstein, David Bensaid, Shaked Brody, Roy Ganz, and Ron Kimmel.
\newblock Fusecap: Leveraging large language models to fuse visual data into enriched image captions.
\newblock \emph{arXiv preprint arXiv:2305.17718}, 2023.

\bibitem[Team et~al.(2024)Team, Mesnard, Hardin, Dadashi, Bhupatiraju, Pathak, Sifre, Rivi{\`e}re, Kale, Love, et~al.]{team2024gemma}
Gemma Team, Thomas Mesnard, Cassidy Hardin, Robert Dadashi, Surya Bhupatiraju, Shreya Pathak, Laurent Sifre, Morgane Rivi{\`e}re, Mihir~Sanjay Kale, Juliette Love, et~al.
\newblock Gemma: Open models based on gemini research and technology.
\newblock \emph{arXiv preprint arXiv:2403.08295}, 2024.

\bibitem[Touvron et~al.(2023)Touvron, Martin, Stone, Albert, Almahairi, Babaei, Bashlykov, Batra, Bhargava, Bhosale, et~al.]{touvron2023llama}
Hugo Touvron, Louis Martin, Kevin Stone, Peter Albert, Amjad Almahairi, Yasmine Babaei, Nikolay Bashlykov, Soumya Batra, Prajjwal Bhargava, Shruti Bhosale, et~al.
\newblock Llama 2: Open foundation and fine-tuned chat models.
\newblock \emph{arXiv preprint arXiv:2307.09288}, 2023.

\bibitem[Wang et~al.(2023)Wang, Cheng, Zhan, Li, Song, and Liu]{wang2023openchat}
Guan Wang, Sijie Cheng, Xianyuan Zhan, Xiangang Li, Sen Song, and Yang Liu.
\newblock Openchat: Advancing open-source language models with mixed-quality data.
\newblock \emph{arXiv preprint arXiv:2309.11235}, 2023.

\bibitem[Yue et~al.(2024)Yue, Ni, Zhang, Zheng, Liu, Zhang, Stevens, Jiang, Ren, Sun, et~al.]{yue2024mmmu}
Xiang Yue, Yuansheng Ni, Kai Zhang, Tianyu Zheng, Ruoqi Liu, Ge~Zhang, Samuel Stevens, Dongfu Jiang, Weiming Ren, Yuxuan Sun, et~al.
\newblock Mmmu: A massive multi-discipline multimodal understanding and reasoning benchmark for expert agi.
\newblock In \emph{Proceedings of the IEEE/CVF Conference on Computer Vision and Pattern Recognition}, pp.\  9556--9567, 2024.

\bibitem[Zhang et~al.(2023)Zhang, Gong, Zhang, Li, Qiao, Ouyang, and Yue]{zhang2023meta}
Yiyuan Zhang, Kaixiong Gong, Kaipeng Zhang, Hongsheng Li, Yu~Qiao, Wanli Ouyang, and Xiangyu Yue.
\newblock Meta-transformer: A unified framework for multimodal learning.
\newblock \emph{arXiv preprint arXiv:2307.10802}, 2023.

\bibitem[Zheng et~al.(2024)Zheng, Chiang, Sheng, Zhuang, Wu, Zhuang, Lin, Li, Li, Xing, et~al.]{zheng2024judging}
Lianmin Zheng, Wei-Lin Chiang, Ying Sheng, Siyuan Zhuang, Zhanghao Wu, Yonghao Zhuang, Zi~Lin, Zhuohan Li, Dacheng Li, Eric Xing, et~al.
\newblock Judging llm-as-a-judge with mt-bench and chatbot arena.
\newblock \emph{Advances in Neural Information Processing Systems}, 36, 2024.

\bibitem[Zhu et~al.(2023)Zhu, Chen, Shen, Li, and Elhoseiny]{zhu2023minigpt}
Deyao Zhu, Jun Chen, Xiaoqian Shen, Xiang Li, and Mohamed Elhoseiny.
\newblock Minigpt-4: Enhancing vision-language understanding with advanced large language models.
\newblock \emph{arXiv preprint arXiv:2304.10592}, 2023.

\end{thebibliography}
\bibliographystyle{iclr2025_conference}

\clearpage
\appendix

\section{Appendix}
\subsection{VisCon-100K Properties}
\label{sec:dataprop} 

This section presents various data properties of the VisCon-100K dataset.

\begin{figure}[h]
  \centering
  \begin{subfigure}[b]{0.49\textwidth}
    \includegraphics[width=1\columnwidth]{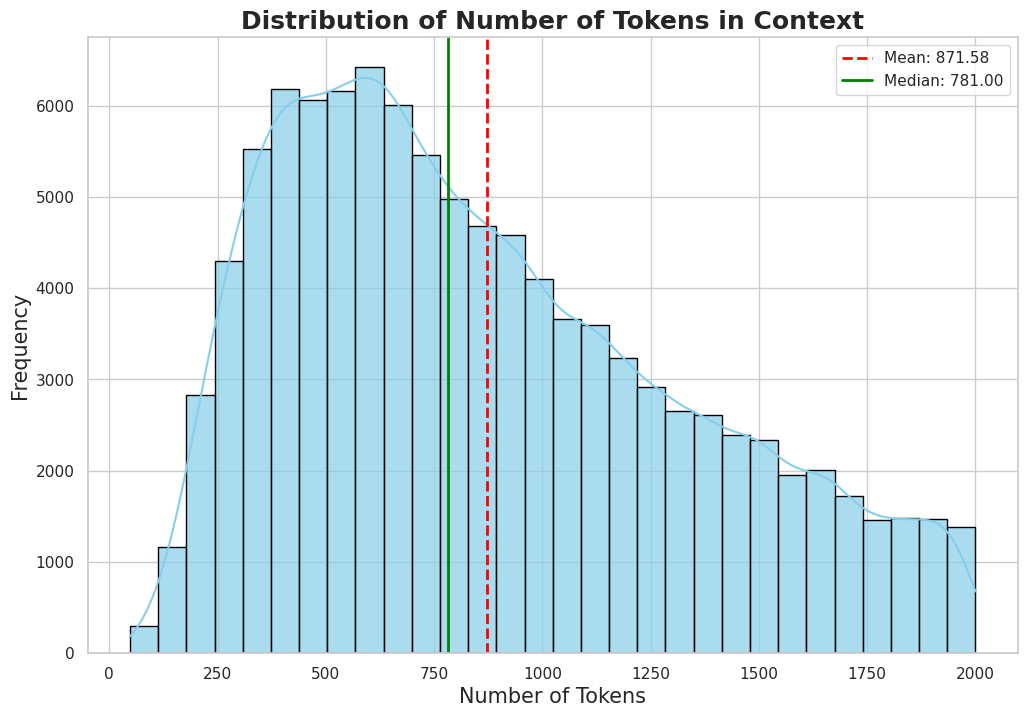}
    \caption{Distribution of Number of Tokens in the Source Context.}
    \label{fig:dist-tokens}
  \end{subfigure}
  \hfill
  \begin{subfigure}[b]{0.49\textwidth}
    \includegraphics[width=1\columnwidth]{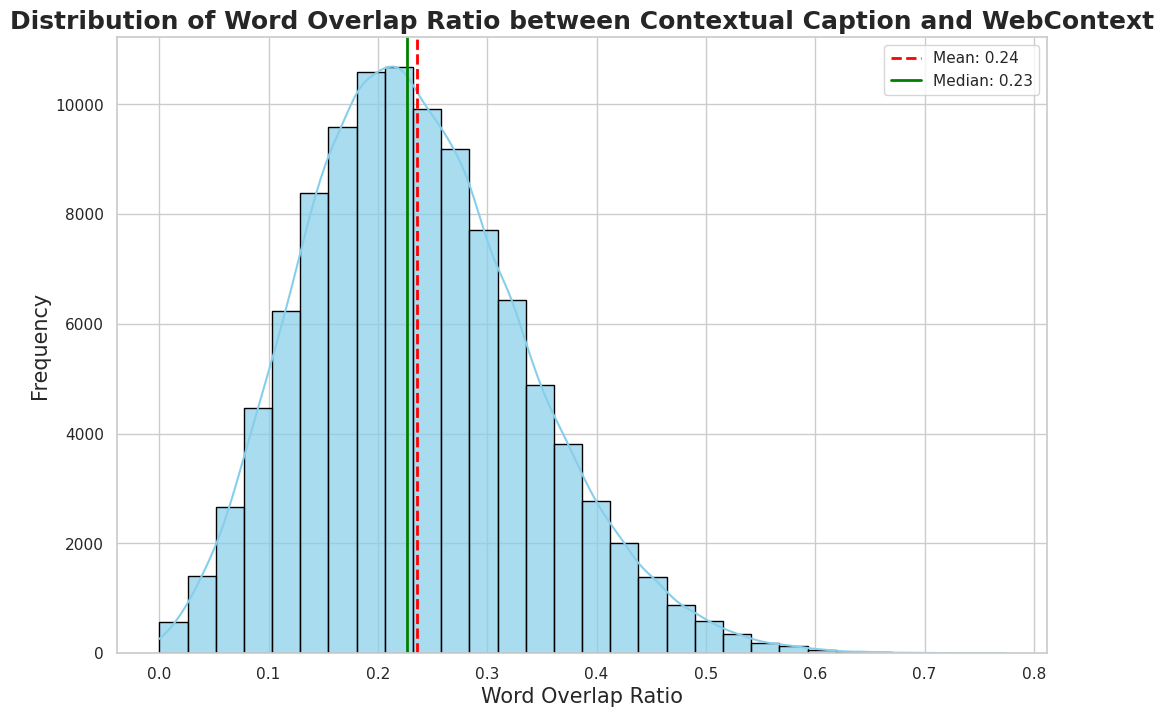}
    \caption{Distribution of Word Overlap Ratio between Contextual Caption and Source Context}
    \label{fig:word-overlap-ratio}
  \end{subfigure}
  \caption{Textual Characteristics of Source Context and their transformed Contextual Captions}
\end{figure}


In Figure~\ref{fig:dist-tokens}, the histogram illustrates that most web documents have a token count between 500 and 1000, indicating a substantial amount of context for generating rich image captions. The mean and median values suggest a slightly skewed distribution, with a long tail extending towards higher token counts.


Figure~\ref{fig:word-overlap-ratio} shows the distribution of overlap ratio between contextual caption and the source context which is calculated after removing stopwords and stemming, and  normalized by caption length.
The average overlap ratio of 0.24 demonstrates the utility of VisCon-100K in augmenting image descriptions with relevant contextual information.

\begin{figure}[h]
  \centering
  \begin{subfigure}[b]{0.49\textwidth}
    \includegraphics[width=\linewidth]{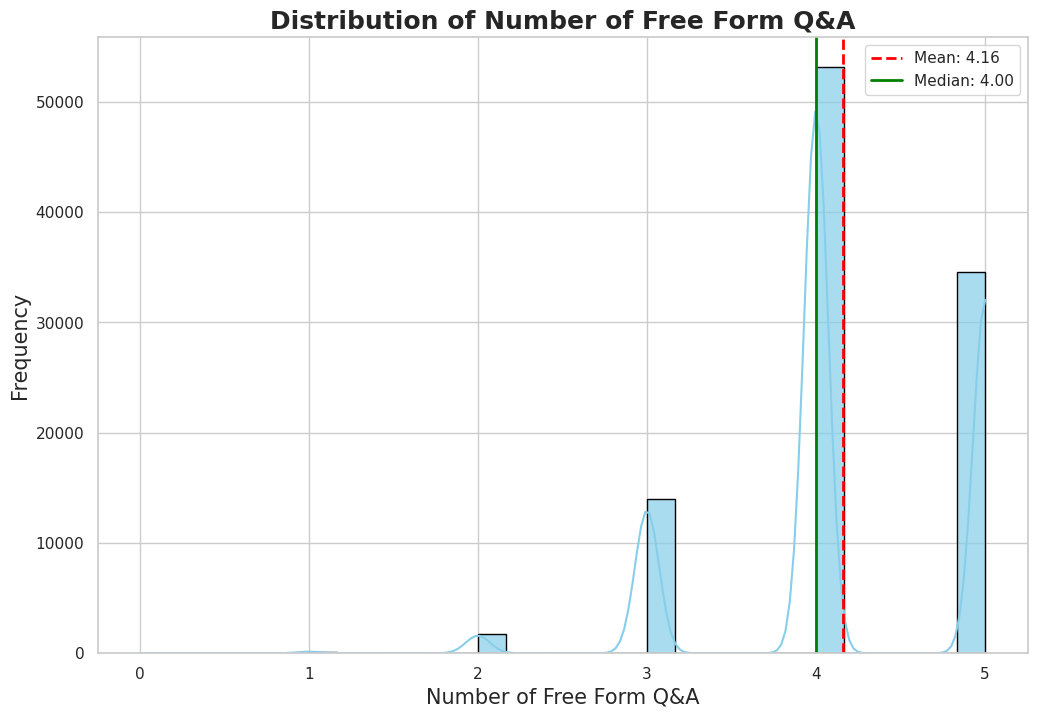}
    \caption{Distribution of Number of Free Form Q\&A}
    \label{fig:dist-ff-qa}
  \end{subfigure}
  \hfill
  \begin{subfigure}[b]{0.49\textwidth}
    \includegraphics[width=\linewidth]{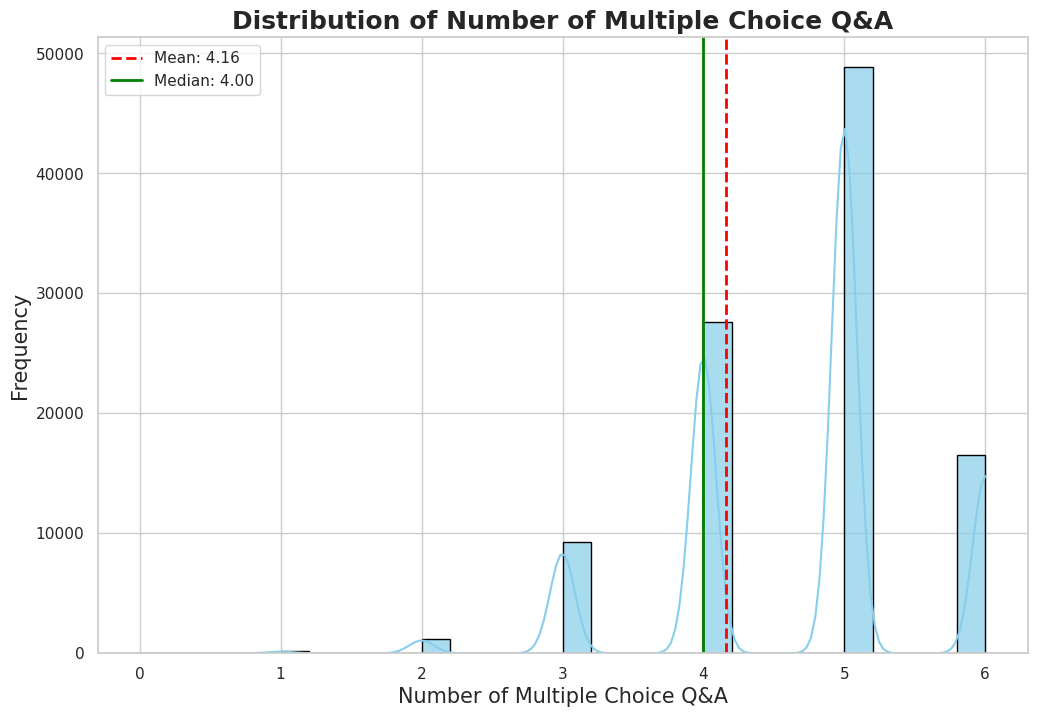}
    \caption{Distribution of Number of Multiple Choice Q\&A}
    \label{fig:dist-mc-qa}
  \end{subfigure}
  \caption{Distributions of Q\&A Types}
\end{figure}


The plot in Figure~\ref{fig:dist-ff-qa} shows that the majority of samples contain 4 free-form Q\&A pairs, which aligns with the dataset's design to provide detailed conversational data.


Figure~\ref{fig:dist-mc-qa} illustrates most samples also contain 4 multiple-choice Q\&A pairs. The similar distribution patterns between free-form and multiple-choice Q\&A pairs facilitate a balanced training approach, allowing models to handle both types of queries effectively.

\begin{figure}[!h]
  \centering
  \includegraphics[width=0.8\columnwidth]{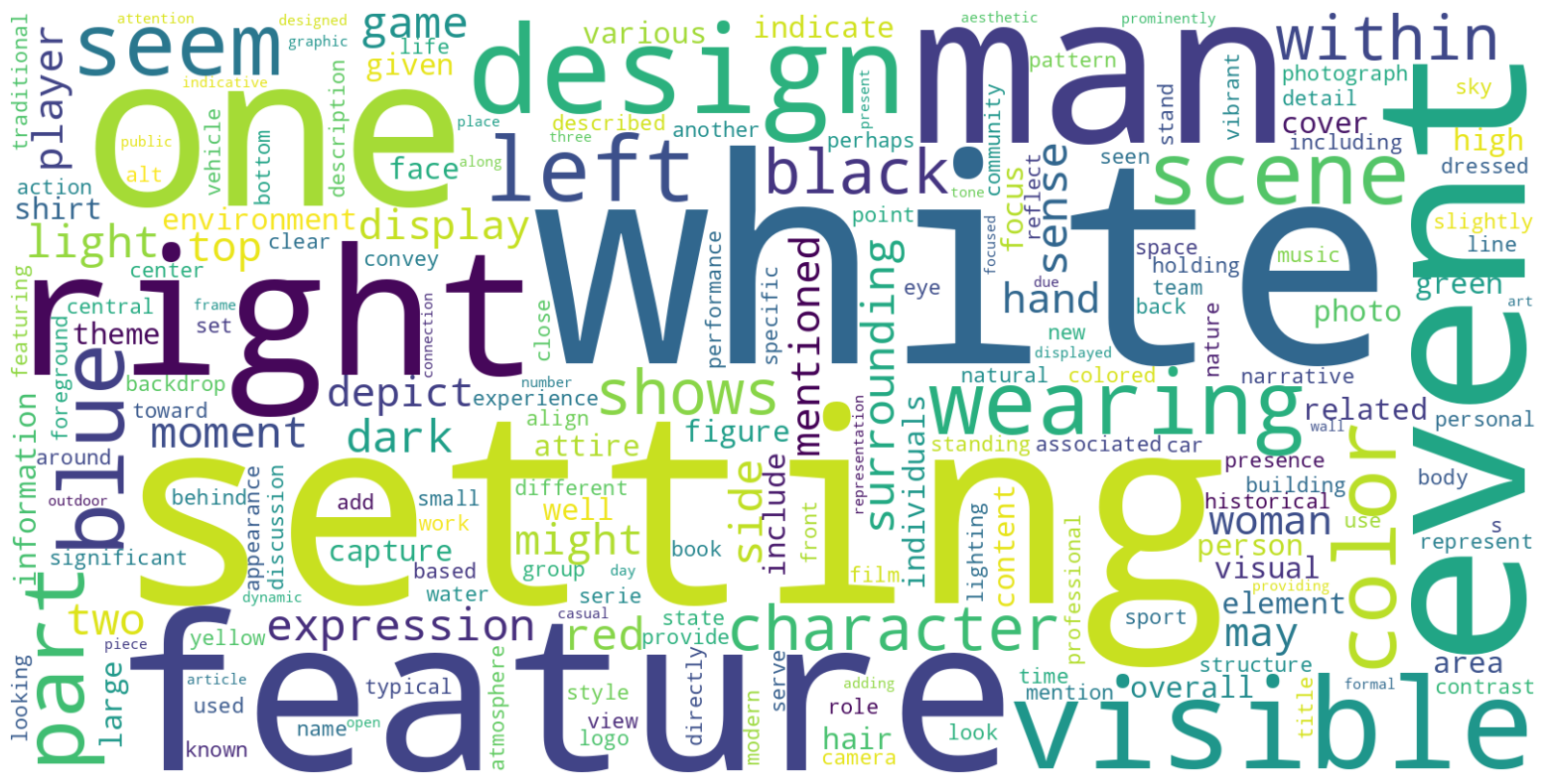}
  \caption{Word Cloud of Captions.}
  \label{fig:word-cloud}
\end{figure}

Figure~\ref{fig:word-cloud} highlights frequently occurring terms such as "one," "white," "right," and "scene," reflecting the common descriptive elements in the dataset's image captions. The prominence of specific terms suggests a focus on detailed visual descriptions, which is critical for enhancing visual understanding in VLMs.

Few examples of the VisCon-100K dataset in Figure~\ref{fig:examples} demonstrate how contextual information from the web pages is used to enhance image descriptions and Q\&A pairs, providing a comprehensive understanding of each image.



\begin{table*}[!h]
\caption{GPT-4 prompt template used to generate contextual captions for images.}
\label{tab:gpt4_prompt}

\centering
\begin{tabular}{|p{0.95\textwidth}|}
\hline
\texttt{Describe the image in detail.} \\ \\

\texttt{Additionally, use the webpage's contextual information along with the alt-text provided below to enrich the description. Understand the webpage information based on its domain name. Focus on the text surrounding the <image> tag, which denotes the input image, and consider other images mentioned as <another-image>. Use only the webpage information relevant to the input image and strictly ignore any information that is not present in the input image. Strictly do not mention the webpage source in the description.} \\ \\ 

\texttt{Webpage URL: \{webpage\_url\}} \\ 

\texttt{Alt-Text: \{alt\_text\}} \\ 

\texttt{Webpage Context: \{webpage\_context\}}
\\ 
\hline
\end{tabular}
\end{table*}

\begin{table*}[!h]
\caption{Prompt template used to convert contextual captions to free-form Q\&A pairs.}
\label{tab:qa_prompt_ffq}

\centering
\begin{tabular}{|p{0.95\textwidth}|}
\hline

\texttt{\#\#\# Human: } \\ \\

\texttt{You are an AI visual assistant, and you are seeing a single image. You are provided with the detailed description of the same image you are looking at. Answer all questions as you are seeing the image.} \\ \\

\texttt{Design a conversation between you and a person asking about this photo. Strictly use `<Human>` and `<Assistant>` as identifiers and the conversation must have only 3 to 5 rounds. The answers should be in a tone that a visual AI assistant is seeing the image and answering the question. Ask diverse questions and give corresponding answers.} \\ \\ 

\texttt{Include questions asking about the visual and the contextual content found in the image description. The visual content covers the object types, counting the objects, object actions, object locations, relative positions between objects, etc. Only include questions that have definite answers:
(1) one can see the content in the image that the question asks about and can answer confidently;
(2) one can determine confidently from the image that it is not in the image. Do not ask any question that cannot be answered confidently.} \\  \\

\texttt{Also include complex questions that are relevant to the content in the image, for example, asking about background knowledge of the objects in the image, asking to discuss about events happening in the image, etc. Again, do not ask about uncertain details. Provide detailed answers when answering complex questions. For example, give detailed examples or reasoning steps to make the content more convincing and well-organized. You can include multiple paragraphs if necessary.} \\ \\

\texttt{Remember to not output more than 5 rounds.} \\ \\

\texttt{\{few\_shot\_examples\}} \\ \\ 

\texttt{Image Description:} \\ \\

\texttt{\{text\}} \\ \\

\texttt{\#\#\# Assistant:}
\\ 
\hline
\end{tabular}
\end{table*}

\begin{table*}[!h]
\caption{Prompt template used to convert contextual captions to multiple-choice Q\&A pairs.}
\label{tab:qa_prompt_mcq}

\centering
\begin{tabular}{|p{0.95\textwidth}|}
\hline

\texttt{\#\#\# Human: } \\ \\

\texttt{You are an AI visual assistant, and you are seeing a single image. You are provided with the detailed description of the same image you are looking at. Answer all questions as you are seeing the image.} \\ \\

\texttt{Design a set of multiple choice questions between you and a person asking about this photo.
Strictly provide 4 choices A., B., C. and D. where only one is valid. and Strictly use `<Human>`,`<Options>` and `<Assistant>` as identifiers for the question, options (new line character delimited) and answer (include only the letter option), and the conversation must have only 3 to 5 rounds. The answers should be in a tone that a visual AI assistant is seeing the image and answering the question. Ask diverse contextual questions and give corresponding answers. Additionally, questions should be independent from each others.} \\ \\ 

\texttt{Include questions asking about the visual and the contextual content found in the image description. The visual content covers the object types, counting the objects, object actions, object locations, relative positions between objects, etc. Only include questions that have definite answers:
(1) one can see the content in the image that the question asks about and can answer confidently;
(2) one can determine confidently from the image that it is not in the image. Do not ask any question that cannot be answered confidently.} \\  \\

\texttt{Also include complex questions that are relevant to the content in the image, for example, asking about background knowledge of the objects in the image, asking to discuss about events happening in the image, etc. Again, do not ask about uncertain details. Provide detailed answers when answering complex questions. For example, give detailed examples or reasoning steps to make the content more convincing and well-organized. You can include multiple paragraphs if necessary.} \\ \\

\texttt{Remember to not output more than 5 rounds.} \\ \\

\texttt{\{few\_shot\_examples\}} \\ \\ 

\texttt{Image Description:} \\ \\

\texttt{\{text\}} \\ \\

\texttt{\#\#\# Assistant:}
\\ 
\hline
\end{tabular}
\end{table*}

\end{document}